\documentclass{article}

\usepackage[preprint]{neurips_2025}

\usepackage[utf8]{inputenc} 
\usepackage[T1]{fontenc}    
\usepackage{hyperref}       
\usepackage{url}            
\usepackage{booktabs}       
\usepackage{amsfonts}       
\usepackage{nicefrac}       
\usepackage{microtype}      
\usepackage{xcolor}         

\usepackage{amsmath}
\usepackage{amssymb}    
\usepackage{amsthm}     
\usepackage{tcolorbox} 
\usepackage{latexsym}
\usepackage{times}
\usepackage{algorithm}    
\usepackage[noend]{algpseudocode} 
\usepackage{float}        
\usepackage{wrapfig}
\usepackage{lipsum} 
\usepackage{subcaption}
\usepackage[font=small,labelfont=bf]{caption}

\usepackage{inconsolata}

\usepackage{multirow}
\usepackage{wrapfig}
\usepackage{natbib}
\usepackage{multicol}
\usepackage{tabularx}
\usepackage{graphicx}
\usepackage{pifont}
\usepackage{colortbl}
\usepackage{longtable}
\usepackage{ulem}
\usepackage{listings}
\usepackage{xspace}
\usepackage{fancyvrb}
\usepackage{adjustbox}
\usepackage{tikz}
\usepackage{forest}

\newcommand{\ourmodel}[0]{{\textsc{PromptCoT-Mamba}}\xspace}
\newcommand{\ourmathmodel}[0]{{\textsc{PromptCoT-Mamba-Math}}\xspace}

\title{Scaling Reasoning without Attention}

\author{
\textbf{Xueliang Zhao}$^\spadesuit$$^{\bigstar}$\thanks{\hspace{2mm}This work was done during an internship at Ant Group.} \quad
\textbf{Wei Wu}$^{\bigstar}$\thanks{\hspace{2mm}Corresponding authors.} \quad
\textbf{Lingpeng Kong}$^\spadesuit$\footnotemark[2] \\
  $^\spadesuit$The University of Hong Kong \quad
  $^\bigstar$Ant Group \\
\texttt{\{xlzhao,lpk\}@cs.hku.hk}\\
\texttt{wuwei19850318@gmail.com} \\
}

\begin{document}
\maketitle

\begin{abstract}
Large language models (LLMs) have made significant advances in complex reasoning tasks, yet they remain bottlenecked by two core challenges: architectural inefficiency due to reliance on Transformers, and a lack of structured fine-tuning for high-difficulty domains. We introduce \ourmodel, an attention-free language model that addresses both issues through architectural and data-centric innovations. Built on the state space dual (SSD) layers of Mamba-2, our model eliminates the need for self-attention and key-value caching, enabling fixed-memory, constant-time inference. To train it for complex reasoning, we propose a two-phase curriculum fine-tuning strategy based on the \textsc{PromptCoT} synthesis paradigm, which generates pedagogically structured problems via abstract concept selection and rationale-guided generation. On benchmark evaluations, \ourmodel-7B outperforms strong Transformer and hybrid models of comparable scale, and even surpasses the much larger Gemma3-27B by 2.6\% on AIME 24, 0.6\% on AIME 25, and 3.0\% on Livecodebench. These results highlight the potential of state space models as efficient and scalable alternatives to attention-based architectures for high-capacity reasoning.
\end{abstract}

\vspace{-5mm}
\begin{center}
{\scriptsize
\href{https://github.com/inclusionAI/PromptCoT}{\includegraphics[height=1em]{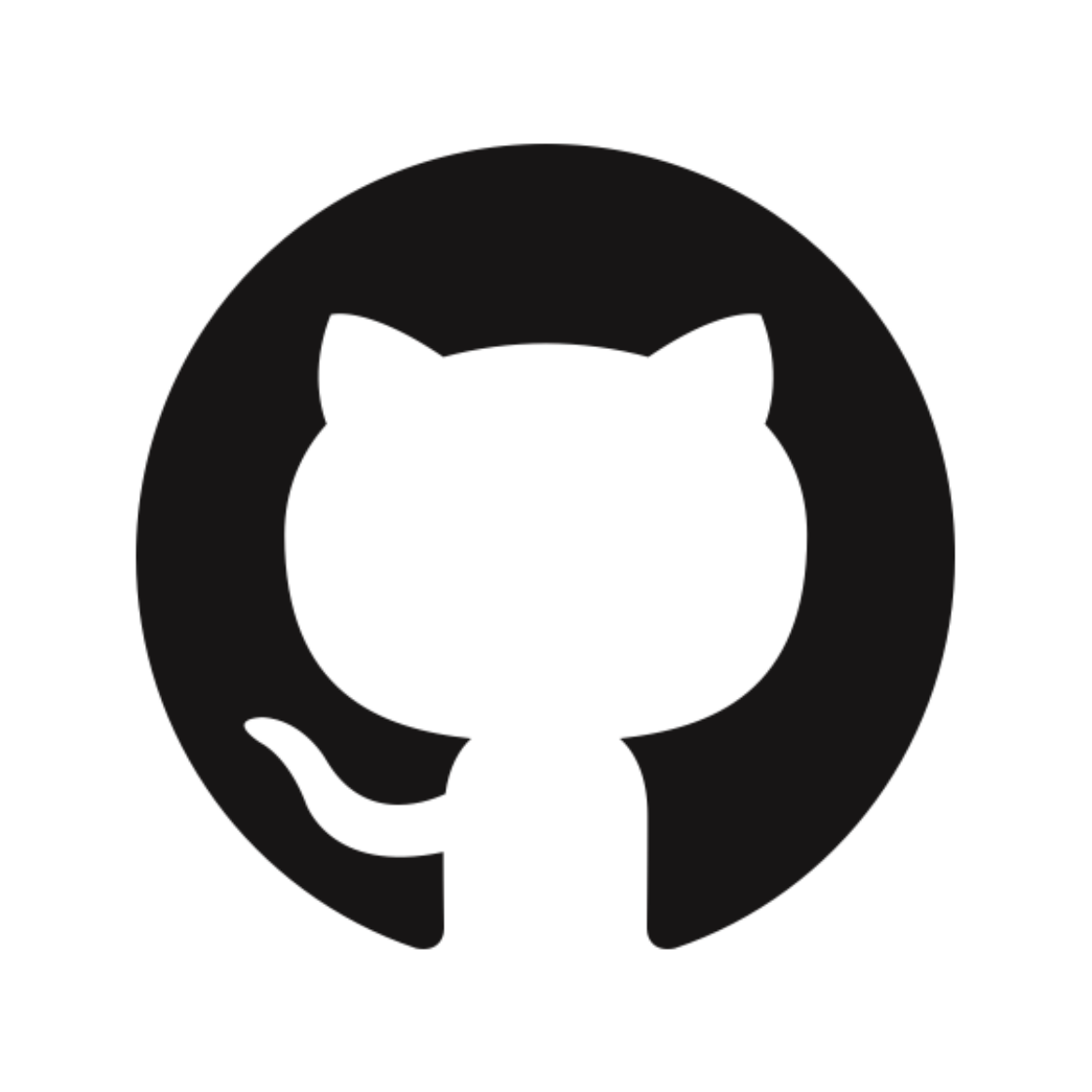} \texttt{github.com/inclusionAI/PromptCoT}} \\
\href{https://huggingface.co/xl-zhao/PromptCoT-Mamba-7B}{\includegraphics[height=1em]{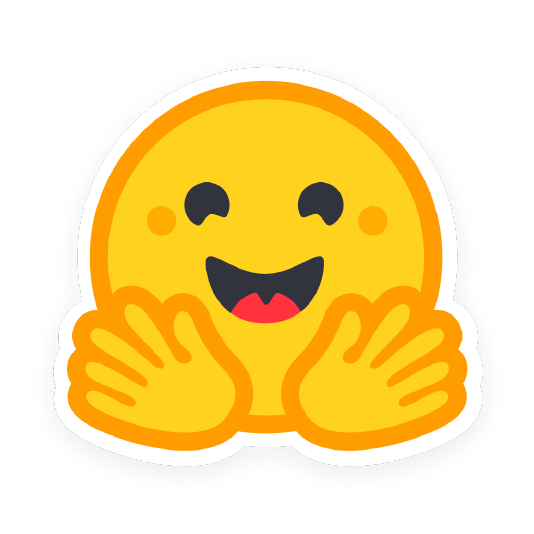} \texttt{huggingface.co/xl-zhao/PromptCoT-Mamba-7B}}
}
\end{center}
\begin{figure}[h]
    \centering
    \includegraphics[width=0.8\linewidth]{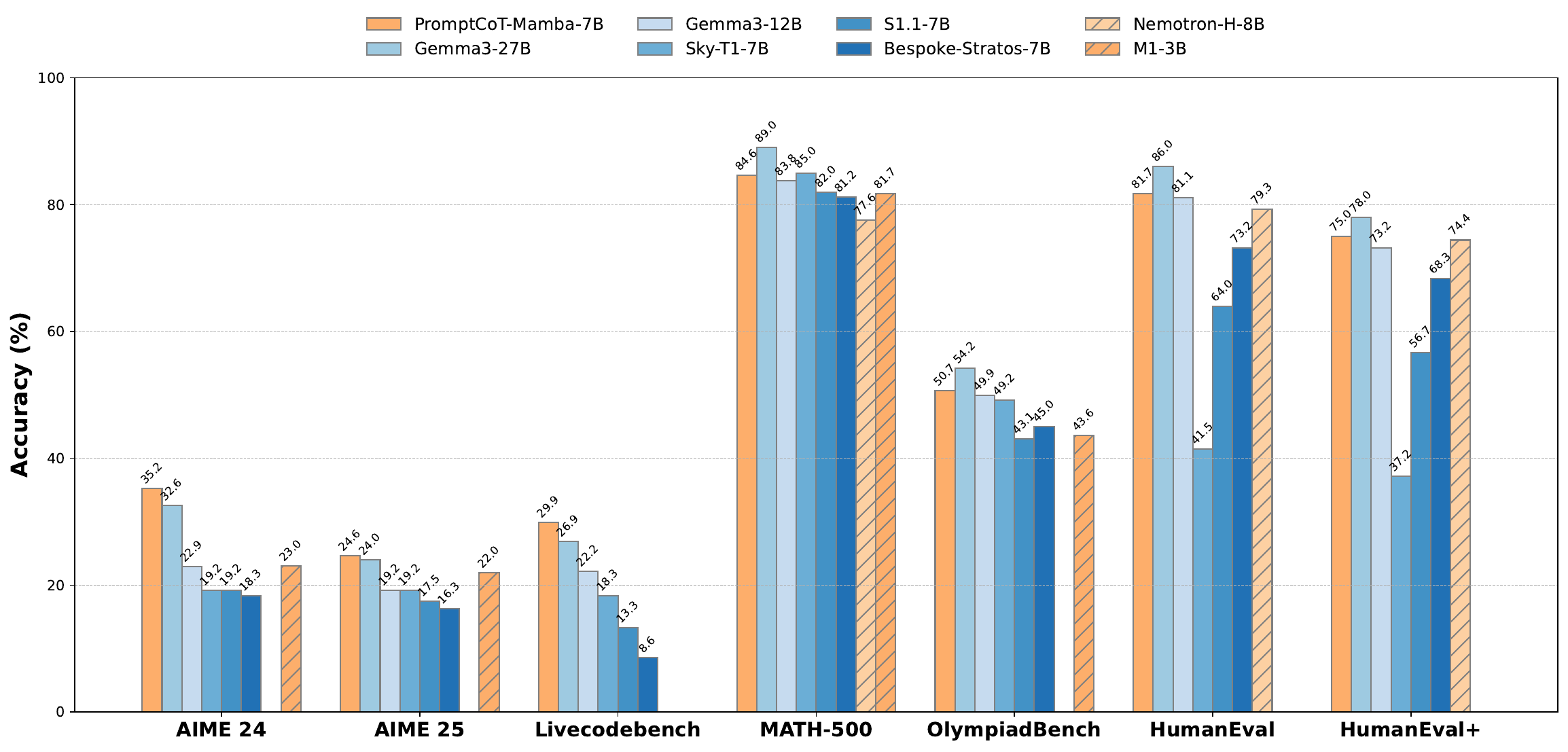}
    \caption{
    Comparison of benchmark performance across models of varying architectures. \ourmodel-7B, a pure attention-free Mamba model (orange), outperforms all Transformer (blue) and hybrid Mamba-Transformer (hatched) baselines of similar or larger scale on AIME 24, AIME 25, and Livecodebench, while remaining competitive across other math and code reasoning benchmarks. Bars represent \textit{pass@1} accuracy.
    }
    \label{fig:model_performance}
\end{figure}

\section{Introduction}
Recent progress in large language models (LLMs) has enabled impressive capabilities in multi-step reasoning~\citep{brown2020language,achiam2023gpt}, mathematical problem solving~\citep{lewkowycz2022solving,lightman2023let}, and code generation\citep{chen2021evaluating}. Despite these advances, a well-known limitation persists in the Transformer architecture that powers most state-of-the-art LLMs: their attention mechanism is memory-intensive and scales poorly with long input contexts. Transformer decoders require maintaining a growing key-value (KV) cache during autoregressive inference, resulting in linear memory growth and degraded throughput in long-context applications. This limitation has become increasingly problematic with the rise of long chain-of-thought reasoning paradigms popularized by models like OpenAI's O1~\citep{jaech2024openai} and DeepSeek's R1~\citep{guo2025deepseek}, where extended reasoning paths are critical for complex problem solving. In this work, we propose \ourmodel, an attention-free language model that delivers constant-time inference and fixed memory per token without maintaining a KV cache, yet outperforms strong Transformer and hybrid models of comparable scale—even surpassing the much larger Gemma3-27B \citep{team2025gemma} by 2.6\% on AIME 24, 0.6\% on AIME 25, and 3.0\% on Livecodebench. To achieve this, \ourmodel builds upon Mamba-2's selective state space architecture~\citep{Dao2024mamba2}, replacing traditional self-attention with state space dual (SSD) layers. This architectural choice delivers constant-time inference and fixed memory consumption while maintaining strong reasoning capabilities. We further enhance the model's problem-solving abilities through a carefully designed two-phase curriculum fine-tuning framework based on the \textsc{PromptCoT} synthesis paradigm~\citep{zhao2025promptcot}. This methodology systematically develops advanced reasoning skills by generating pedagogically structured problems through a principled process: first sampling abstract concepts and constructing expert-style rationales, then generating target problems conditioned on these structured rationales.

This integration of architectural efficiency with structured supervision enables \ourmodel to solve complex reasoning tasks effectively, while offering favorable inference characteristics compared to Transformer-based alternatives. Extensive experiments across math and code benchmarks show that our model outperforms strong open-source baselines of similar size, and even surpasses significantly larger Transformer models on high-difficulty domains such as AIME and Livecodebench.

\noindent
\textbf{Contributions.} This paper makes the following key contributions:
\begin{itemize}
    \item We introduce \ourmodel, a fully attention-free language model that outperforms strong Transformer and hybrid baselines of comparable size—for instance, surpassing S1.1-7B by \textbf{16.0\%} on AIME 24, \textbf{7.1\%} on AIME 25, and \textbf{16.6\%} on LiveCodeBench—establishing a new performance bar for pure Mamba-based architectures.
    \item We propose a scalable two-phase curriculum fine-tuning pipeline based on the \textsc{PromptCoT} synthesis paradigm, which enables the model to acquire expert-level reasoning capabilities through pedagogically structured supervision.
    \item We demonstrate that the math-specialized variant \ourmathmodel achieves state-of-the-art results on all math benchmarks, including \textbf{+3.4\%} on MATH-500, \textbf{+7.7\%} on AIME 24, and \textbf{+6.2\%} on AIME 25 over the general \ourmodel, illustrating the effectiveness of domain-focused adaptation.
    \item We show that \ourmodel delivers up to \textbf{3.66$\times$} higher throughput on 24GB memory and \textbf{1.69$\times$} on 72GB memory compared to strong Transformer baselines, confirming its suitability for efficient long-context inference in resource-constrained settings.
\end{itemize}

\section{Method}

We introduce \ourmodel, an attention-free language model that combines efficient state space modeling with curriculum fine-tuning based on the \textsc{PromptCoT} synthesis paradigm. The architecture is built on Mamba-2~\citep{Dao2024mamba2}, employing SSD layers as the core computational primitive in place of self-attention. We first describe the SSD layer formulation for both autoregressive inference and parallel training~(\S\ref{sec:ssd}), and then describe a two-phase curriculum strategy that scaffolds learning from foundational to advanced reasoning, using PromptCoT-guided synthesis to introduce high-complexity examples~(\S\ref{sec:curriculum}).

\subsection{State Space Dual Layers in Mamba-2}
\label{sec:ssd}

We adopt the SSD layer introduced by \citet{Dao2024mamba2}, which reformulates selective state-space models as recurrent architectures through a structured matrix representation. This section details both inference and training for a single SSD layer, which forms the core computational unit in our model. Let $N$ denote the hidden state dimension, $P$ the intermediate projection dimension, and $d$ the input embedding dimension. The full architecture comprises multiple SSD layers composed sequentially, each maintaining its own hidden state trajectory.

\paragraph{Inference.}

During autoregressive inference, the SSD layer maintains and updates a hidden state recurrently to compute the next token distribution, conditioned on the current token embedding. Let \( w_t \in \mathcal{V} \) denote the input token at position \( t \), and let \( \mathbf{e}_t \in \mathbb{R}^d \) denote its corresponding embedding, where $d$ is the embedding dimension. A shared projection generates all time-dependent parameters:
\[
[a_t, \mathbf{b}_t, \mathbf{c}_t, \mathbf{u}_t] = \textit{Linear}(\mathbf{e}_t),
\]
where \( a_t \in [0, 1] \) is a scalar decay factor, \( \mathbf{b}_t, \mathbf{c}_t \in \mathbb{R}^N \) are the input and output kernel vectors, and \( \mathbf{u}_t \in \mathbb{R}^P \) is a projected feature vector.

The core recurrence consists of a low-rank state update followed by a linear readout:
\begin{align}
\label{eq:original_recurrence}
\mathbf{Z}_t &= \mathbf{u}_t \otimes \mathbf{b}_t \notag \\
\mathbf{H}_t &= a_t \cdot \mathbf{H}_{t-1} + \mathbf{Z}_t  \\
\mathbf{y}_t &= \mathbf{c}_t^\top \mathbf{H}_t^\top \notag
\end{align}
Here, \( \mathbf{Z}_t \in \mathbb{R}^{P \times N} \) represents a low-rank input-dependent update to the hidden state, and \( \mathbf{H}_t \in \mathbb{R}^{P \times N} \) accumulates contributions over time via exponential decay. The output vector \( \mathbf{y}_t \in \mathbb{R}^P \) is produced by linearly projecting the hidden state with the output kernel \( \mathbf{c}_t \).

The output \( \mathbf{y}_t \) is mapped to vocabulary logits using a learned projection head, and the next token is sampled as \( w_{t+1} \sim \mathrm{Categorical}(\mathrm{softmax}(\boldsymbol{\ell}_t)) \), where \( \boldsymbol{\ell}_t = \textit{VocabProj}(\mathbf{y}_t) \).

The per-step complexity of this procedure is $\mathcal{O}(N P)$, owing to the outer product in the state update and the projection in the output computation. Unlike Transformer decoders, which require $\mathcal{O}(T N)$ operations per layer and maintain a growing key-value cache, SSD layers operate with fixed memory and constant computation per token.

\paragraph{Training.}

During training, the SSD recurrence admits a parallel formulation using a series of structured contraction operations. Let $\mathbf{E} \in \mathbb{R}^{T \times d}$ denote the sequence of input embeddings, where $T$ is the sequence length. A shared linear projection produces the time-dependent parameters:
\[
[\mathbf{a}, \mathbf{B}, \mathbf{C}, \mathbf{U}] = \textit{Linear}(\mathbf{E}),
\]
where \( \mathbf{a} \in \mathbb{R}^{T} \) are decay scalars, \( \mathbf{B}, \mathbf{C} \in \mathbb{R}^{T \times N} \) are input and output kernels, and \( \mathbf{U} \in \mathbb{R}^{T \times P} \) are projected features.

We express the SSD training computation using a sequence of structured contractions over time. Let \( S \) be a symbolic contraction axis with \( S = T \). To simplify notation, we denote rank-3 tensors such as \( \mathbf{Z} \) and \( \mathbf{H} \) using bold uppercase letters, with dimensionality made explicit through context.
The forward computation consists of the following steps:
\begin{align}
\mathbf{Z} &= \text{contract}(\mathrm{S\ P},\ \mathrm{S\ N} \rightarrow \mathrm{S\ P\ N})(\mathbf{U}, \mathbf{B}) \notag \\
\mathbf{H} &= \text{contract}(\mathrm{T\ S},\ \mathrm{S\ P\ N} \rightarrow \mathrm{T\ P\ N})(\mathbf{M}, \mathbf{Z}) \\
\mathbf{Y} &= \text{contract}(\mathrm{T\ N},\ \mathrm{T\ P\ N} \rightarrow \mathrm{T\ P})(\mathbf{C}, \mathbf{H}) \notag 
\end{align}

Here, \( \mathbf{Z} \in \mathbb{R}^{T \times P \times N} \) represents the low-rank update tensor, \( \mathbf{H} \in \mathbb{R}^{T \times P \times N} \) is the accumulated hidden state, and \( \mathbf{Y} \in \mathbb{R}^{T \times P} \) contains the output representations at each time step.

The structured transition matrix \( \mathbf{M} \in \mathbb{R}^{T \times T} \) encodes the recurrence dynamics, with entries defined by
\[
m_{j,i} =
\begin{cases}
\prod_{k=i+1}^{j} \mathbf{a}_k & \text{if } i \leq j, \\
0 & \text{otherwise},
\end{cases}
\]
capturing the cumulative decay between positions \( i \) and \( j \).

This formulation is algebraically equivalent to the original recurrence (Eq.~\ref{eq:original_recurrence}) while enabling efficient parallel execution using structured matrix operations. The total training-time complexity of a single SSD layer is \( \mathcal{O}(T N P) \), which is substantially more efficient than the \( \mathcal{O}(T^2 N) \) cost of Transformer self-attention, especially in long-context scenarios.

\subsection{Curriculum Fine-Tuning with \textsc{PromptCoT}}
\label{sec:curriculum}

To facilitate the acquisition of complex reasoning abilities, we adopt a two-phase curriculum learning strategy that progresses from simpler to more challenging objectives. This framework follows established principles in curriculum learning, in which training data is staged by complexity to improve optimization stability and efficiency.

We initialize the training pipeline from Mamba Codestral~\citep{mistral2024codestral}, and organize fine-tuning into an \textit{initialization phase} followed by an \textit{advanced phase}.
In the initialization phase, we continue fine-tuning the model on open-source datasets focused on fundamental reasoning tasks. Specifically, we employ \textsc{OpenCodeReasoning}~\citep{ahmad2025opencodereasoning} and \textsc{OpenThoughts2}~\citep{openthoughts}, which combine a small number of expert-authored problems from competitions such as Codeforces and the AIME mathematics contest, alongside a large collection of automatically synthesized examples. 
The majority of training instances are generated using automatic synthesis pipelines based on \textsc{NuminaMath}~\citep{li2024numinamath} and \textsc{Mammoth}~\citep{yue2023mammoth,yue2024mammoth2}, which produce high-quality examples spanning diverse reasoning domains.

In the advanced phase, the core of our training strategy centers on data synthesized using \textsc{PromptCoT}~\citep{zhao2025promptcot}. Although PromptCoT was originally developed for Olympiad-level mathematical problem generation, we demonstrate that its methodology can be seamlessly adapted to other domains, including programmatic reasoning. PromptCoT formulates task synthesis as a two-stage, rationale-guided generation process: given a set of foundational concepts, it first generates a pedagogical rationale that reflects expert-level abstractions over concept composition and problem construction strategies. The final problem is then constructed conditioned on both the rationale and the concept set.

PromptCoT jointly optimizes the likelihood of rationale generation and problem generation, resulting in examples that reflect expert-level abstraction and deductive depth. In our adaptation, we customize the prompt design and concept selection pipeline to produce challenging and pedagogically structured data for both mathematical and code-based tasks. This stage provides the most demanding problems in the training corpus and plays a central role in developing high-capacity reasoning behaviors.

\section{Experiments}
\subsection{Benchmarks}
We evaluate our models across seven diverse benchmarks spanning mathematics and code generation. Each dataset targets a different aspect of reasoning or problem-solving:

\textbf{(1) MATH-500}~\citep{lightman2023let, hendrycks2021measuring} consists of 500 curated high-school competition problems from the original MATH dataset, assessing advanced mathematical reasoning and problem-solving skills. Problems span algebra, geometry, combinatorics, and number theory.

\textbf{(2,3) AIME 24 and 25}~\citep{aimedata} are annual high-school mathematics competitions from the American Mathematics Competitions (AMC) series. We use the latest publicly available AIME 24 and 25 problems, evaluating models under the \texttt{avg@16} metric by sampling 16 generations per problem, following standard practices in prior work.

\textbf{(4) OlympiadBench}~\citep{he2024olympiadbench} is a recently introduced benchmark containing 8,476 Olympiad-level mathematics and physics problems. We evaluate only on the mathematical subset and focus on the English problems, following the official evaluation protocol. 

\textbf{(5) HumanEval}~\citep{chen2021evaluating} is a widely-used code synthesis benchmark that requires models to generate Python functions from docstrings. Each problem is evaluated by executing test cases to verify functional correctness.

\textbf{(6) HumanEval+}~\citep{liu2023your} extends HumanEval with an 80$\times$ larger test suite per problem, exposing subtle bugs and testing deeper behavioral correctness. It reveals systematic overestimation in previous evaluation pipelines and improves robustness.

\textbf{(7) LiveCodeBench (v5)}~\citep{jain2024livecodebench} is a live-updated benchmark that sources real-world coding problems from LeetCode, AtCoder, and CodeForces released between May 2023 and May 2024. To ensure contamination-free evaluation, we use the v5 split and report \texttt{avg@8}, sampling 8 completions per problem following the benchmark's protocol.

\subsection{Evaluation Metrics}
We adopt \textit{pass@1} accuracy as the primary metric across all datasets. For math benchmarks, an output is considered correct if it exactly matches the ground-truth final boxed answer. For code synthesis tasks (HumanEval, HumanEval+, LiveCodeBench), correctness is determined by passing all functional unit tests provided in the benchmark.

For AIME 24 and 25, and LiveCodeBench, we report \texttt{avg@k} accuracy by sampling \(k = 16\) and \(k = 8\) generations respectively and averaging over the per-problem correctness. This protocol accounts for the inherent sampling variance in open-ended generation and aligns with practices from prior literature.

\subsection{Baselines}
We compare our approach against a set of competitive open-source models spanning both standard Transformer and hybrid Mamba-Transformer architectures. These baselines represent the current frontier in reasoning and code generation.

\textbf{(1) Sky-T1-7B}~\citep{sky_t1_2025} is a Transformer-based instruction-tuned model developed by NovaSky. It adopts a standard decoder-only Transformer architecture and is optimized for multi-step reasoning tasks.

\textbf{(2) S1.1-7B}~\citep{muennighoff2025s1} builds on Qwen2.5-7B and applies a targeted finetuning strategy to improve reasoning depth. The model retains a full Transformer architecture.

\textbf{(3) Bespoke-Stratos-7B}~\citep{bespoke_stratos} is a Transformer model trained using distillation from larger teacher models. It emphasizes long-form reasoning performance through curated training pipelines.

\textbf{(4,5) Gemma3-12B and Gemma3-27B}~\citep{team2025gemma} are instruction-tuned Transformer models developed by Google DeepMind. They employ alternating local and global attention layers to support efficient long-context processing.

\textbf{(6) Nemotron-H-8B}~\citep{blakeman2025nemotron} adopts a \textit{hybrid Mamba-Transformer} architecture, replacing most attention layers with Mamba-2 blocks while preserving a small number of attention layers for global context modeling. This design improves efficiency while maintaining strong accuracy.

\textbf{(7) M1-3B}~\citep{wang2025m1} is a compact \textit{hybrid Mamba-Transformer} model trained via cross-architecture distillation. It achieves strong performance under compute-efficient conditions.

\subsection{Implementation Details}

All experiments are conducted on a single node equipped with 8×A100 80GB GPUs using DeepSpeed ZeRO Stage 2 for distributed training and memory optimization. We adopt a two-stage curriculum learning approach where the initialization stage employs 1.88M prompt-completion pairs, comprising 735k samples from OpenCodeReasoning and 1.14M samples from OpenThoughts2, with completions generated using state-of-the-art reasoning models (Deepseek-R1~\citep{guo2025deepseek} and QwQ~\citep{qwq32b}) to ensure high-quality reasoning traces. The advanced stage incorporates 256k prompt-completion pairs constructed via the \textsc{PromptCoT} pipeline, augmented with 246k from OpenCodeReasoning and 232k from OpenMathReasoning to balance the difficulty of the training curriculum. Both stages utilize the AdamW optimizer with a learning rate of $5 \times 10^{-6}$, $\beta_1 = 0.9$, $\beta_2 = 0.95$, and weight decay of $0.01$, with warmup for the first 100 steps followed by a fixed learning rate schedule and a global batch size of 64. The initialization stage processes sequences of maximum length 16,384 tokens, while the advanced stage extends this to 20,480 tokens to accommodate longer reasoning chains. During evaluation, we set the maximum generation length to 65,536 tokens to allow complete reasoning trace generation.

\begin{table}[t]
\centering
\caption{
Evaluation results on seven benchmarks. All values are reported as \textit{pass@1} accuracy. For AIME 24 and AIME 25, we report averaged pass@1 over 16 samples per problem (avg@16); for Livecodebench-v5, we report averaged pass@1 over 8 samples (avg@8). The best and second-best results are shown in \textbf{bold} and \underline{underline}, respectively. Transformer models are shaded in blue; hybrid Mamba-Transformer models are shaded in orange.
}
\label{tab:main_results}
\resizebox{0.95\textwidth}{!}{
\begin{tabular}{lccccccc}
\toprule
\textbf{Model} & \textbf{MATH-500} & \textbf{AIME 24} & \textbf{AIME 25} & \textbf{OlympiadBench} & \textbf{HumanEval} & \textbf{HumanEval+} & \textbf{Livecodebench} \\
\midrule
\rowcolor{blue!4}
Sky-T1-7B & 85.0 & 19.2 & 19.2 & 49.2 & 41.5 & 37.2 & 18.3 \\
\rowcolor{blue!4}
S1.1-7B & 82.0 & 19.2 & 17.5 & 43.1 & 64.0 & 56.7 & 13.3 \\
\rowcolor{blue!4}
Bespoke-Stratos-7B & 81.2 & 18.3 & 16.3 & 45.0 & 73.2 & 68.3 & 8.6 \\
\rowcolor{blue!4}
Gemma3-12B & 83.8 & 22.9 & 19.2 & 49.9 & 81.1 & 73.2 & 22.2 \\
\rowcolor{blue!4}
Gemma3-27B & \textbf{89.0} & \underline{32.6} & \underline{24.0} & \textbf{54.2} & \textbf{86.0} & \textbf{78.0} & \underline{26.9} \\
\rowcolor{orange!5}
Nemotron-H-8B & 77.6 & -- & -- & -- & 79.3 & 74.4 & -- \\
\rowcolor{orange!5}
M1-3B & 81.7 & 23.0 & 22.0 & 43.6 & -- & -- & -- \\
\rowcolor{gray!10}
\ourmodel-7B & \underline{84.6} & \textbf{35.2} & \textbf{24.6} & \underline{50.7} & \underline{81.7} & \underline{75.0} & \textbf{29.9} \\
\bottomrule
\end{tabular}
}
\end{table}

\subsection{Main Results}

The results in Table~\ref{tab:main_results} demonstrate the effectiveness of our proposed \ourmodel-7B across diverse reasoning and code generation tasks. (1) \ourmodel-7B outperforms all Transformer and hybrid Mamba-Transformer baselines with comparable parameter sizes, marking the first instance where a pure Mamba-based architecture exceeds the performance of similarly sized attention-based models across this range of benchmarks. This underscores the strength of structured state space modeling when paired with curriculum-oriented training; (2) In addition, \ourmodel-7B outperforms significantly larger Transformer models, including Gemma3-27B, on several of the most challenging tasks. It achieves new state-of-the-art results on AIME 24 (35.2\%), AIME 25 (24.6\%), and Livecodebench-v5 (29.9\%), demonstrating strong reasoning and generalization; (3) These results reveal the growing potential of pure Mamba architectures as competitive and scalable alternatives to Transformer-based models. With careful training and data design, Mamba models can match or exceed the performance of larger attention-based systems, offering a compelling direction for efficient sequence modeling in high-complexity domains.

\section{Discussions}
Beyond the comprehensive benchmark comparisons, we further analyze the behavior and characteristics of our method through targeted research questions. Specifically, we aim to answer:  
(1) \textbf{RQ1:} How do different training stages and data sources contribute to overall performance?  
(2) \textbf{RQ2:} Is it possible to strategically reduce code-related capabilities in favor of more specialized and extreme mathematical reasoning performance?  
(3) \textbf{RQ3:} How does the inference efficiency of the Mamba-based architecture compare to standard Transformers? These questions guide our ablation studies, performance trade-off analysis, and efficiency profiling, discussed in the following sections.

\subsection{Ablation Study for RQ1}

\begin{table}[t]
\centering
\caption{
Ablation results on AIME 24, AIME 25, and Livecodebench-v5. ``- PromptCoT'' removes the curriculum synthesis stage. ``- PromptCoT \& OT'' uses only OpenCodeReasoning (OCR). ``- PromptCoT \& OCR'' uses only OpenThoughts (OT). All values are reported as avg@16 for AIME and avg@8 for Livecodebench.
}
\label{tab:ablation}
\resizebox{0.55\textwidth}{!}{
\begin{tabular}{lccc}
\toprule
\textbf{Model} & \textbf{AIME 24} & \textbf{AIME 25} & \textbf{Livecodebench} \\
\midrule
Full & \textbf{35.2} & \textbf{24.6} & \textbf{29.9} \\ \midrule
- PromptCoT & 11.7 & 8.3 & 5.3 \\
- PromptCoT \& OT & 1.7 & 3.3 & 13.6 \\
- PromptCoT \& OCR & 19.2 & 13.3 & 12.5 \\
\bottomrule
\end{tabular}
}
\end{table}

To evaluate the contribution of each training phase, we conduct an ablation study on AIME 24, AIME 25, and Livecodebench-v5, as shown in Table~\ref{tab:ablation}. We examine three variants of our method: (1) exclusion of the PromptCoT synthesis phase (``- PromptCoT''), which retains only the foundational finetuning stage using OpenThoughts and OpenCodeReasoning; (2) removal of both PromptCoT and OpenThoughts (``- PromptCoT \& OT''), finetuning the model solely on OpenCodeReasoning; and (3) removal of both PromptCoT and OpenCodeReasoning (``- PromptCoT \& OCR''), using only OpenThoughts.

The results show that the full \ourmodel-7B achieves the best performance across all benchmarks, underscoring the importance of curriculum-driven synthesis for high-complexity tasks. Removing PromptCoT causes substantial degradation, particularly on AIME 24/25, highlighting its role in teaching abstract multi-step reasoning. When both PromptCoT and OpenThoughts are removed, performance drops even further, revealing that OpenThoughts contributes harder, more diverse samples compared to OpenCodeReasoning. Conversely, removing OpenCodeReasoning but retaining OpenThoughts leads to modest recovery, confirming that both sources of supervision are complementary. Overall, the curriculum structure, progressing from basic to advanced supervision, is critical for effective reasoning generalization.

\subsection{Mathematical Specialization for RQ2}

\begin{table}[t]
\centering
\caption{
Comparison between the math-specialized (\ourmathmodel) and general (\ourmodel) variants. 
}
\label{tab:math_vs_general}
\resizebox{0.95\textwidth}{!}{
\begin{tabular}{lccccccc}
\toprule
\textbf{Model} & \textbf{MATH-500} & \textbf{AIME 24} & \textbf{AIME 25} & \textbf{OlympiadBench} & \textbf{HumanEval} & \textbf{HumanEval+} & \textbf{Livecodebench} \\
\midrule
PromptCoT-Mamba-Math-7B & \textbf{88.0} & \textbf{42.9} & \textbf{30.8} & \textbf{52.1} & 71.3 & 66.5 & 20.3 \\
PromptCoT-Mamba-7B      & 84.6 & 35.2 & 24.6 & 50.7 & \textbf{81.7} & \textbf{75.0} & \textbf{29.9} \\
\bottomrule
\end{tabular}
}
\end{table}

To investigate whether domain specialization improves mathematical reasoning, we train a variant of our model, \ourmathmodel-7B, with a focus solely on math. Specifically, we replace OpenCodeReasoning and OpenThoughts in the initialization stage with \textsc{OpenMathReasoning}~\citep{moshkov2025aimo}, a curated corpus of mathematical tasks designed to provide high-quality foundational supervision.

As shown in Table~\ref{tab:math_vs_general}, this math-specialized model achieves significant gains on all math benchmarks: 88.0\% on MATH-500, 42.9\% on AIME 24, 30.8\% on AIME 25, and 52.1\% on OlympiadBench. These improvements confirm that narrowing the pretraining distribution allows the model to better internalize domain-specific patterns and abstractions.
However, this specialization comes at the expense of generalization. On code-oriented tasks such as HumanEval (71.3\%), HumanEval+ (66.5\%), and Livecodebench-v5 (20.3\%), the performance lags behind the full \ourmodel-7B. This result illustrates a clear trade-off between general-purpose coverage and task-specific optimization, and highlights the flexibility of the Mamba architecture for targeted domain adaptation.

\subsection{Inference Efficiency for RQ3}

\begin{figure}[H]
    \centering
    \begin{subfigure}[b]{0.48\textwidth}
        \centering
        \includegraphics[width=\textwidth]{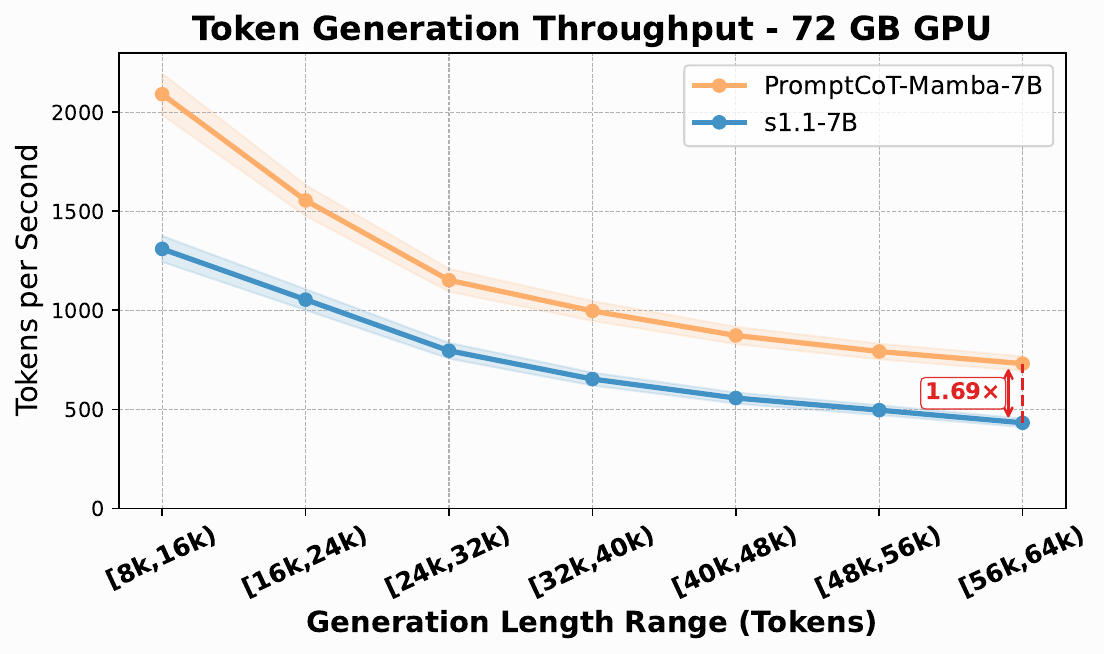}
        \caption{72 GB GPU}
    \end{subfigure}
    \hfill
    \begin{subfigure}[b]{0.48\textwidth}
        \centering
        \includegraphics[width=\textwidth]{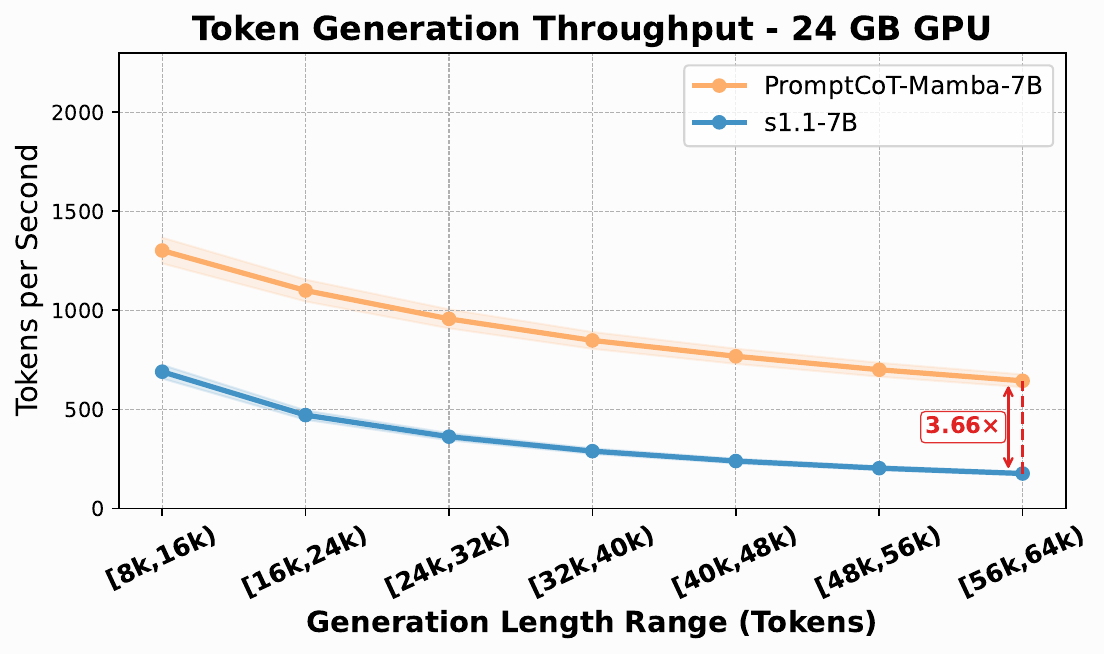}
        \caption{24 GB GPU}
    \end{subfigure}
    \caption{
    Token generation throughput (tokens/sec) of \ourmodel-7B and \textsc{s1.1-7B} under different GPU memory configurations. Performance is measured across a range of generation lengths.
    }
    \label{fig:throughput_comparison}
\end{figure}

We compare the inference efficiency of \ourmodel-7B against a strong Transformer baseline, s1.1-7B~\citep{muennighoff2025s1}, using \texttt{vLLM 0.7.3} on a single NVIDIA A100 80GB GPU. For each generation length interval \([a, b)\), we set \texttt{min\_tokens} = \(a\) and \texttt{max\_tokens} = \(b\) to mimic variable-length decoding workloads commonly encountered in practical deployments.

Figure~\ref{fig:throughput_comparison} reports token generation throughput (tokens per second) across two hardware memory configurations: 24GB and 72GB. We restrict the accessible GPU memory accordingly to simulate inference scenarios at different deployment tiers. \ourmodel-7B achieves consistently higher throughput across all lengths. Under the 24GB setting, it yields a \textbf{3.66×} improvement over s1.1-7B at the longest sequence range. Even under 72GB, where the Transformer baseline benefits from greater caching capacity, our model still achieves a \textbf{1.69×} speedup. These results highlight the practical strengths of our proposed \ourmodel, which enables efficient, high-throughput inference under both memory-constrained and long-context conditions. The combination of architectural efficiency and curriculum-driven training positions it as a compelling alternative to attention-based models for scalable deployment in real-world scenarios.

\section{Related Work}

\subsection{Reasoning with Large Language Models}

The study of reasoning in LLMs has evolved rapidly, driven by advances in both prompting strategies and data-centric methodologies. Early work demonstrated that reasoning behavior can be elicited through carefully designed prompts, such as chain-of-thought prompting~\citep{wei2022chain}, least-to-most decomposition~\citep{zhou2022least}, and reflective self-correction~\citep{shinn2024reflexion}. While these methods reveal the latent capabilities of pretrained models, they often rely on implicit reasoning heuristics and are sensitive to prompt design.
Subsequent efforts have increasingly shifted toward training-time approaches, emphasizing dataset construction and model supervision~\citep{wang2022self,zhao2023sego,zhao2024subgoalxl}. Recent works have introduced open-source corpora curated for high-quality mathematical and programmatic reasoning tasks~\citep{ahmad2025opencodereasoning,moshkov2025aimo}, while synthesis pipelines such as NuminaMath~\citep{li2024numinamath} and Mammoth~\citep{yue2024mammoth2} enable scalable generation of diverse and structured examples. In this context, \textsc{PromptCoT}~\citep{zhao2025promptcot} proposes a rationale-guided synthesis paradigm that produces abstract, compositional examples with expert-style intermediate reasoning. We adopt and extend this formulation across both mathematical and code-based domains in our approach.

\subsection{Linear Sequence Modeling}

A growing body of work explores alternatives to attention mechanisms for improved efficiency in sequence modeling. Linearized attention architectures~\citep{katharopoulos2020transformers,zheng2022linear,zheng2023efficient,yang2024parallelizing}, hardware-friendly designs such as GLA~\citep{yang2023gated}, and recurrence-based models including RWKV~\citep{peng2023rwkv} and RetNet~\citep{sun2023retentive} have demonstrated promising results in language modeling tasks. State-space approaches like Mamba~\citep{gu2023mamba,Dao2024mamba2} further enhance modeling capacity through selective and structured recurrence. 
While these models achieve competitive performance in language modeling, they are not specifically designed for multi-step reasoning. Our work addresses this gap by developing a reasoning-centric framework for linear architectures, demonstrating that such models can match or surpass Transformer baselines on complex mathematical and code generation benchmarks.

\section{Conclusion}

We present \ourmodel, the first attention-free language model to match and exceed the performance of strong Transformer and hybrid baselines on challenging math and code tasks. Built on Mamba-2’s structured state space architecture, it delivers constant-time, fixed-memory inference, addressing key inefficiencies of attention-based models. Central to our approach is a two-phase curriculum fine-tuning pipeline based on the PromptCoT paradigm, which enables the model to learn complex reasoning skills from challenging data. Extensive results show that \ourmodel not only outperforms comparably sized models but also surpasses larger Transformers in accuracy and throughput, especially under constrained or long-context settings. Our findings underscore the viability of attention-free architectures for efficient, high-performance language modeling.

\bibliographystyle{plainnat}
\bibliography{custom}

\end{document}